\documentclass[runningheads]{llncs}
\usepackage[T1]{fontenc}
\usepackage{graphicx}
\usepackage{xcolor}
\usepackage{amsmath,amssymb,bm}
\usepackage{booktabs}
\usepackage{multirow}
\usepackage{placeins}
\usepackage{marvosym}   

\newcommand{\norm}[1]{\left\lVert #1 \right\rVert}
\newcommand{\soft}{\operatorname{soft}}

\begin{document}
\title{Interpretable Hyperspectral Unmixing Framework with Fixed Endmember Prior and Structured Residual Refinement\thanks{This work is accepted to 23rd Pacific Rim
International Conference on Artificial Intelligence (PRICAI 2026), and partly supported by Hunan Provincial Natural Science Foundation of China, under the Science and Technology Innovation Program of Hunan Province (No.~2025JJ60883); the Hunan Provincial College Students' Entrepreneurship Training Program (No.~S202510530147X); and the National College Students' Innovation Training Program (No.~202510530057).}}
\titlerunning{Interpretable Hyperspectral Unmixing with Residual Refinement}
%
\author{Ziyi Guan\inst{1} \and
Jianping Zhang\inst{1}\textsuperscript{\Letter} \and
Qian Liu\inst{1}}
\authorrunning{Z. Guan et al.}
\institute{School of Mathematics and Computational Science, Xiangtan University, Xiangtan 411105, China \\
\email{jpzhang@xtu.edu.cn}}
\maketitle              
\begin{abstract}
Hyperspectral unmixing decomposes mixed pixels into material endmembers and their abundances from contiguous spectral observations. In modular sensing pipelines, endmembers are often first identified and then treated as fixed during abundance estimation. When this fixed endmember prior is inaccurate, spatially structured mismatch arising from illumination changes, sensor artifacts, or material boundaries may be incorrectly captured by the abundance variables, leading to unstable decompositions. This study presents an interpretable stage-wise hyperspectral unmixing framework (I-HyperSU) under fixed endmember priors, which is explicitly decomposed into a fixed endmember matrix $\mathbf{A}$, an abundance block $\mathbf{X}$, and a structural residual refinement block $\mathbf{S}$. The X-block estimates abundances using FISTA with nonnegativity and sparsity enhancement, and a soft penalty that approximately enforces sum-to-one constraints. The S-block jointly applies low-rank SVD structural regularization and a lightweight deep image prior (DIP) to refine structured residuals. This staged design makes the interaction between abundance and residual components transparent and interpretable. Experiments on Samson, Urban, and Jasper Ridge datasets demonstrate that, under fixed and imperfect endmember priors, soft abundance relaxation consistently outperforms hard simplex projection. Under the default N-FINDR endmember prior, the proposed framework reduces the joint reconstruction error by 61.7\%--69.5\% compared with a fixed-$\mathbf{A}$ UCLS baseline, while keeping the abundance RMSE nearly unchanged, indicating that the residual refinement branch accounts for structured model mismatch without degrading the abundance estimates. For example, on Urban, the reconstruction SAM decreases from $5.99^\circ$ for the X-only model to $1.92^\circ$ for the full model.

\end{abstract}
\section{Introduction}
\begin{sloppypar}
Hyperspectral imaging extends visual perception beyond conventional RGB by resolving material-dependent responses across many contiguous spectral bands. Hyperspectral unmixing aims to decompose mixed-pixel spectra into endmember spectra and their corresponding abundances, enabling applications such as material detection, environmental assessment, and urban scene interpretation \cite{bioucas2012overview,zou2025review}. Such a representation is particularly important when limited spatial resolution and complex scene configurations cause individual pixels to contain multiple materials.
\end{sloppypar}

In typical modular pipelines, endmembers are first estimated using geometric methods, spectral libraries, or calibration procedures, and are then provided to subsequent stages as fixed priors for abundance estimation \cite{nascimento2005vca,winter1999nfindr,ren2003atgp,chang2006fippi}. Although this fixed-endmember approach is widely adopted, it introduces a key difficulty: once these priors are set, any spectral mismatch or unmodeled effects must be accommodated exclusively by the downstream unmixing process.

\begin{sloppypar}
Existing methods range from constrained abundance optimization techniques and variability-sensitive formulations to newer deep learning and transformer-based models. Classical constrained and sparse-regression methods, such as FCLS \cite{heinz2001} and SUnSAL \cite{iordache2011}, impose simplex-like abundance constraints \cite{iordache2012tv}. These approaches perform well when the endmembers are specified accurately, but they run the risk of embedding systematic model mismatch into the abundance estimates. Models that are designed to handle variability and mismatch, such as those in \cite{halimi2016,fu2016danser,drumetz2016elmm,thouvenin2016plmm}, explicitly account for spectral mismatch or endmember variability, yet they typically do so by modifying the endmember matrix or estimating it jointly. Recent deep learning methods, such as UnDIP \cite{rasti2022undip} and MAT-Net \cite{wang2024matnet}, exploit deep image priors \cite{ulyanov2018dip} or learned spatial--spectral representations, yet they are primarily designed for joint or learned unmixing rather than strictly modular frameworks that rely on fixed, non-adaptive endmember priors. As a result, they offer no explicit guidance on how a downstream solver should allocate unexplained energy between abundance corrections and structured residual refinement when the prior has to be kept fixed. In contrast, this paper focuses on a frozen-prior setting where the endmember matrix $\mathbf{A}$ is kept fixed, and explicitly separates abundance estimation from per-scene structured residual refinement without using external training data.
\end{sloppypar}

\begin{sloppypar}
To explicitly represent this allocation under imperfect fixed priors, we propose an interpretable stage-wise hyperspectral unmixing framework (I-HyperSU), where $\mathbf{A}$ serves as the fixed endmember prior, $\mathbf{X}$ corresponds to the abundance branch, and $\mathbf{S}$ represents the structural-residual branch. The $\mathbf{X}$-block uses FISTA \cite{beck2009} with nonnegativity constraints, a sparsity-promoting regularizer, and a soft sum-to-one penalty, which reduces the tendency of hard simplex projections to conceal mismatch within the abundances. The $\mathbf{S}$-block uses low-rank structural coordinates to guide a lightweight deep image prior (DIP)--based residual refinement module, modeling structured discrepancies that are not captured by the endmember model. At each iteration, the stage-wise procedure produces intermediate pairs $(\mathbf{X}^k,\mathbf{S}^k)$, so that the progressive reduction of mismatch between abundances and residuals can be directly interpreted.
\end{sloppypar}

The main contributions are summarized as follows.
\begin{itemize}
\item We present a staged hyperspectral unmixing framework with fixed endmember priors, in which the FISTA-based abundance estimation and the refinement of structured residuals are clearly decoupled.
\item We propose a low-rank-guided SVD-DIP branch to capture structured residuals, where low-rank information is integrated with a lightweight nonlinear refinement module, yielding an explicit and controllable mechanism to model spectral mismatches.
\item We perform controlled experiments on the Samson, Urban, and Jasper Ridge datasets using fixed N-FINDR endmember priors, showing that our framework reduces the joint reconstruction error by $61.7\%$--$69.5\%$ compared to the UCLS fixed-prior baseline across all three datasets, while maintaining comparable abundance RMSE. This demonstrates the advantages of soft abundance relaxation and explicit residual modeling.
\item On synthetic data with a known ground-truth residual, we verify that the residual branch captures genuine out-of-model mismatch rather than merely fitting observation residuals, achieving $r\!\approx\!0.77$ overall and $r\!>\!0.99$ on the component orthogonal to the endmember subspace, while remaining inactive when no mismatch is present.
\end{itemize}

The remainder of the paper is structured as follows. Section~\ref{sec:method} formulates the problem with fixed endmember priors and introduces the stage-wise hyperspectral unmixing algorithm. Section~\ref{sec:experiments} reports experimental results, including fixed-prior validation, comparisons between soft- and hard-prior simplex constraints, ablations of the S-block, visualizations of residual refinement, and evaluations against classical baseline methods.

\section{Method}
\label{sec:method}
We begin by formulating hyperspectral unmixing problem using fixed endmember priors, after which we estimate the abundance matrix under a relaxed abundance constraint and explicitly channel the remaining structured mismatch into a separate residual refinement branch. As a result, we obtain an optimization-driven decomposition framework rather than a purely end-to-end trained network.

\subsection{Problem Formulation}
Let $\mathbf{Y}=[\bm{y}_1,\ldots,\bm{y}_n]\in\mathbb{R}^{\ell\times n}$ denote the hyperspectral data matrix observed, where $\ell$ is the number of spectral bands and $n$ is the total number of pixels. The matrix $\mathbf{A}=[\bm{a}_1,\ldots,\bm{a}_r]\in\mathbb{R}^{\ell\times r}$ collects the endmember signatures, and $\mathbf{X}=[\bm{x}_1,\ldots,\bm{x}_n]\in\mathbb{R}^{r\times n}$ contains the corresponding abundance vectors. The standard linear mixing model (LMM) can be expressed as
\begin{equation}
\mathbf{Y} = \mathbf{A}\mathbf{X}, \qquad \mathbf{X} \succeq 0,\qquad \mathbf{1}_r^{\top}\mathbf{X} = \mathbf{1}_n^{\top},
\label{eq:ideal}
\end{equation}
where $\mathbf{1}_r \in \mathbb{R}^{r}$ and $\mathbf{1}_n \in \mathbb{R}^{n}$ denote all-ones vectors. These conditions ensure that each abundance vector $\bm{x}_i$ belongs to the probability simplex
\begin{equation}
\Delta = \{\bm{x} \in \mathbb{R}^{r} \mid \bm{x} \succeq 0,\ \mathbf{1}_r^{\top}\bm{x} = 1\}.
\label{eq:simplex}
\end{equation}

In practice, the endmember matrix $\mathbf{A}$ is typically obtained from a geometric endmember extraction algorithm, and then kept fixed during the subsequent unmixing stage. When $\mathbf{A}$ is inaccurate, the observations cannot always be faithfully modeled by $\mathbf{A}\mathbf{X}$ alone. Real scenes may exhibit shadows, striping artifacts, background structures, and other systematic effects that are not captured by the fixed endmember model. To address these, we consider
\begin{equation}
\mathbf{Y} = \mathbf{A}\mathbf{X} + \mathbf{S} + \bm{\eta},
\label{eq:obs}
\end{equation}
where $\mathbf{S}\in\mathbb{R}^{\ell\times n}$ represents a structured residual component and $\bm{\eta}$ denotes small random perturbations. From this perspective, a general decomposition can be written as
\begin{equation}
\min_{\mathbf{X},\mathbf{S}}\;
\frac{1}{2}\norm{\mathbf{Y}-\mathbf{A}\mathbf{X}-\mathbf{S}}_F^2
\;+\;\lambda_\mathbf{X} \psi_\mathbf{X}(\mathbf{X})
\;+\;\lambda_\mathbf{S} \psi_\mathbf{S}(\mathbf{S}),
\label{eq:generic_joint}
\end{equation}
where $\psi_\mathbf{X}(\cdot)$ and $\psi_\mathbf{S}(\cdot)$ are regularizers imposed on the abundance and residual terms, respectively. Thus, the main focus of this paper is to determine how the unmodeled signal should be apportioned between $\mathbf{X}$ and $\mathbf{S}$ once $\mathbf{A}$ is fixed.

\subsection{Proposed I-HyperSU Framework}
\begin{figure}[!tbp]
\centering
\includegraphics[width=0.92\textwidth]{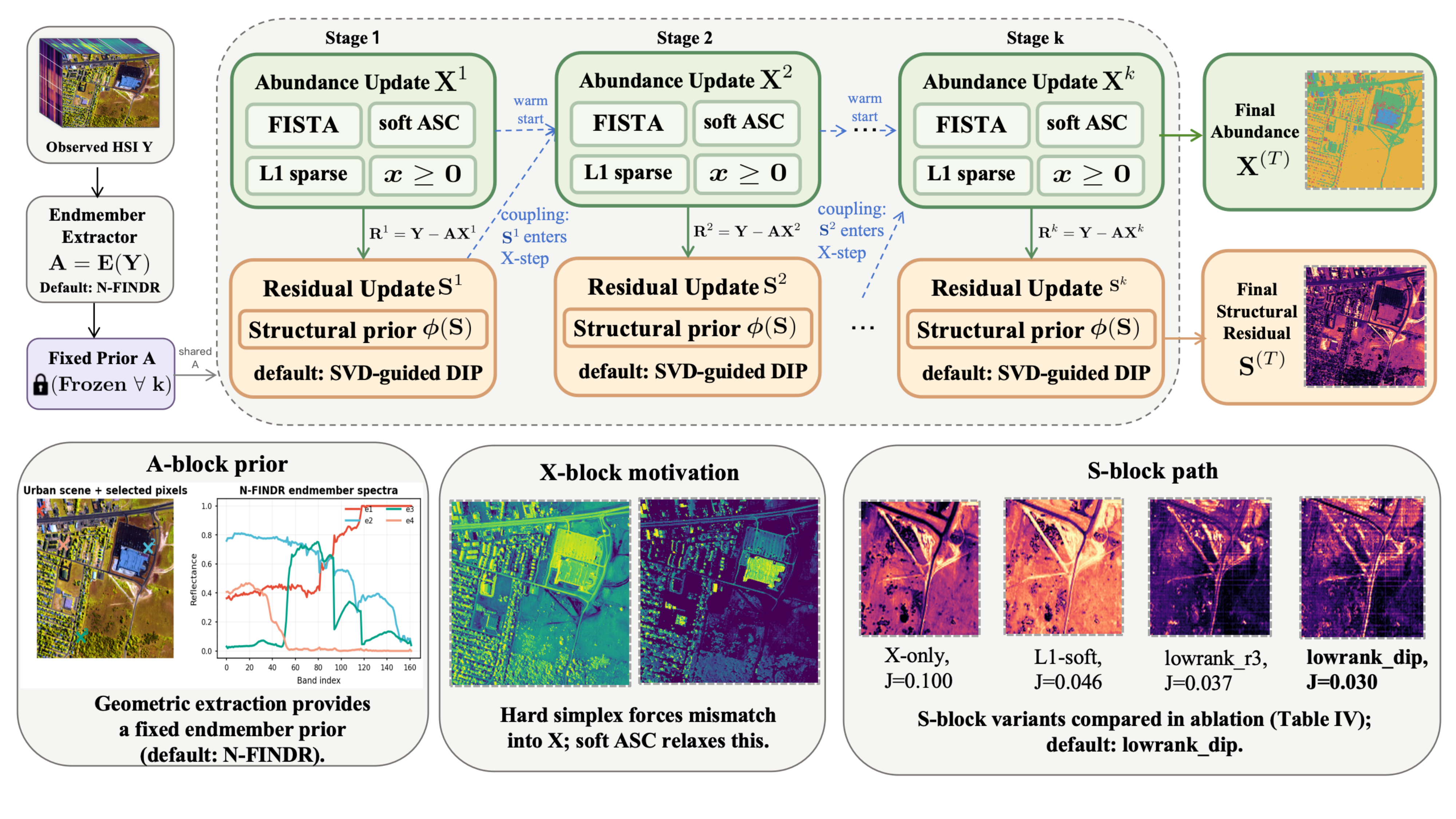}
\caption{Overview of the proposed interpretable stage-wise hyperspectral unmixing framework (I-HyperSU). The A-block provides a frozen endmember prior, the X-block performs abundance estimation with soft relaxation, and the S-block captures structured mismatch through progressive residual refinement.}
\label{fig:axs_framework}
\end{figure}

Fig.~\ref{fig:axs_framework} provides an overview of the proposed interpretable stage-wise hyperspectral unmixing framework. The A-block provides a fixed geometric prior extracted from the observed HSI; the X-block estimates abundance maps under relaxed physical constraints, incorporating $\ell_1$ sparsity and nonnegativity alongside a soft ASC penalty that replaces the hard simplex projection to reduce mismatch absorption; and the S-block captures structured mismatch through a progression of models ranging from element-wise soft thresholding (\texttt{l1-soft}) to low-rank truncated SVD (\texttt{lowrank\_r3}) and DIP-guided low-rank decomposition (\texttt{lowrank\_dip}).

For any observation $\mathbf{Y}$, an endmember extractor $\mathcal{E}(\cdot)$ first produces a frozen prior
$\mathbf{A} = \mathcal{E}(\mathbf{Y})$, which is fixed throughout all subsequent optimization.
The remaining decomposition proceeds across $T$ stages.
At stage $k$, the X-block takes the previous residual estimate $\mathbf{S}^{k-1}$ as coupled input
and solves the abundance map $\mathbf{X}^k$ using FISTA with nonnegativity and a
soft sum-to-one penalty; and the S-block fits a structural residual $\mathbf{S}^k \approx \mathbf{R}^k=\mathbf{Y} - \mathbf{A}\mathbf{X}^k$
using low-rank structural guidance and a lightweight DIP model.
The abundance estimate $\mathbf{X}^k$ warm-starts the next X-block, while $\mathbf{S}^k$ re-enters as a coupled residual input to the following X-step, which is formally written as
\begin{equation}
  (\mathbf{X}^k, \mathbf{S}^k) = \mathcal{T}(\mathbf{X}^{k-1}, \mathbf{S}^{k-1}; \mathbf{A}).
  \label{eq:stage_recursion}
\end{equation}
After the $T$ stages, the final output is the abundance map $\mathbf{X}^T$
and the structural residual $\mathbf{S}^T$.

\subsubsection{Fixed Endmember Extraction ($\mathbf{A}$-Block).}
As discussed, the endmember matrix $\mathbf{A}$ is estimated once before the main optimization and then held fixed. Within the geometric framework of unmixing \cite{heinz2001,bioucas2012overview}, the A-block imposes a fixed endmember prior by selecting a small subset of pixels that approximate the vertices of the data simplex:
\begin{equation}
\mathbf{A} = \mathcal{E}(\mathbf{Y}),
\end{equation}
where $\mathcal{E}(\cdot)$ denotes a geometric endmember extractor.

Different algorithms implement $\mathcal{E}$ using different geometric rules. We adopt four classical pure-pixel strategies for the A-block: VCA \cite{nascimento2005vca}, N-FINDR \cite{winter1999nfindr}, ATGP \cite{ren2003atgp}, and FIPPI \cite{chang2006fippi}. All are designed to pick pixels that are spectrally extreme or highly pure, and the selection of pixels is given as
\begin{equation}
\Omega^{\star}
=\mathcal{I}_{\mathcal{E}}(\mathbf{Y},r)=
\arg\max_{\Omega,\ |\Omega|=r}
\operatorname{Vol}(\mathbf{Y}_{\Omega}),
\end{equation}
where $\operatorname{Vol}(\mathbf{Y}_{\Omega})$ is the volume of the simplex spanned by the selected spectra, and $\Omega=\{\omega_1,\ldots,\omega_r\}\subset\{1,\ldots,N\}$.

The A-block itself is not the core algorithmic contribution; its purpose is merely to provide a fixed geometric prior to the hyperspectral unmixing solver. Unless stated otherwise, we use N-FINDR as the default endmember extractor:
\begin{equation}
\Omega_{\mathrm{NF}}
=
\mathcal{I}_{\mathrm{N\text{-}FINDR}}(\mathbf{Y},r),
\qquad
\mathbf{A} = \mathbf{Y}_{\Omega_{\mathrm{NF}}}.
\end{equation}
For every dataset, we perform endmember extraction separately, producing a dataset-specific fixed prior. VCA, ATGP, and FIPPI are employed solely as alternative choices for $\mathcal{E}$ in our comparisons and are neither merged with nor used to refine the N-FINDR-based prior.

Once $\mathbf{A}$ is fixed, we restrict the decomposition problem to the case of an imperfect yet fixed prior, and we explicitly specify how the model mismatch is apportioned between the abundance matrix $\mathbf{X}$ and the residual term $\mathbf{S}$.

\subsubsection{Soft Abundance Relaxation ($\mathbf{X}$-Block).}
Under ideal conditions, each abundance vector $\bm{x}_i$ belongs to the simplex $\Delta$, which enforces both nonnegativity and a unit-sum constraint. When $\mathbf{A}$ is accurately known, the projection onto this simplex has a clear geometric interpretation. However, the estimated endmember matrix $\mathbf{A}$ is generally imperfect, so strictly enforcing simplex membership compels the solution to remain in $\Delta$ regardless of the reconstruction quality, which can inadvertently shift structured model mismatches into the abundance variables.

To tackle this, we adopt a soft relaxation of the abundance constraint:
\begin{equation}
\mathbf{X}^k = \arg\min_{\mathbf{X}}\; g(\mathbf{X})+ \lambda_x \norm{\mathbf{X}}_1+ \iota_{\mathbb{R}_+}(\mathbf{X}),
\label{eq:x_branch}
\end{equation}
\begin{sloppypar}
\noindent where
$g(\mathbf{X})=
\frac{1}{2}\norm{\mathbf{Y}-\mathbf{A}\mathbf{X}-\mathbf{S}^{k-1}}_F^2
+\frac{\mu}{2}\norm{\mathbf{1}^{\top}\mathbf{X}-\mathbf{1}^{\top}}_F^2$,
$\iota_{\mathbb{R}_+}(\mathbf{X})$ enforces the nonnegativity of $\mathbf{X}$, $\norm{\mathbf{X}}_1$ encourages sparsity~\cite{iordache2011}, and the soft sum-to-one penalty $\norm{\mathbf{1}^{\top}\mathbf{X}-\mathbf{1}^{\top}}_F^2$ permits controlled deviations from the simplex constraint. Consequently, the abundance variables are no longer required to account for all mismatches induced by fixed priors, and these structured discrepancies are instead explicitly modeled within the S-block.
\end{sloppypar}

To solve \eqref{eq:x_branch} with FISTA~\cite{beck2009}, we first evaluate the gradient as
\begin{equation}
\nabla g(\mathbf{X})=
\mathbf{A}^{\top}(\mathbf{A}\mathbf{X}+\mathbf{S}^{k-1}-\mathbf{Y})
+\mu\,\mathbf{1}
\left(\mathbf{1}^{\top}\mathbf{X}-\mathbf{1}^{\top}\right).
\label{eq:grad_g}
\end{equation}
Next, the step size \(\alpha=1/L_g\) is approximated using the Lipschitz constant \(L_g=\lambda_{\max}(\mathbf{A}^\top \mathbf{A} + \mu\,\mathbf{1}\mathbf{1}^\top)\) of \(\nabla g\).

For each outer iteration \(k\), we run an inner loop indexed by \(s\). To keep the notation concise, the stage index is dropped within this inner loop. The initialization is given by
\begin{equation}
\mathbf{X}^{0}=\mathbf{X}^{k-1},\qquad \hat{\mathbf{X}}^{0}=\mathbf{X}^{k-1},\qquad q_0=1.
\end{equation}
 The FISTA updates then take the form
\begin{align}
\widetilde{\mathbf{X}}^{s+1} &= \mathbf{X}^{s}-\alpha\nabla g(\mathbf{X}^{s}),
& \hat{\mathbf{X}}^{s+1} &= \max\!\left(0,\soft_{\alpha\lambda_x}(\widetilde{\mathbf{X}}^{s+1})\right),\\
q_{s+1} &=(1+\sqrt{1+4q_s^2})/{2},
& \mathbf{X}^{s+1} &= \hat{\mathbf{X}}^{s+1}+\tfrac{q_s-1}{q_{s+1}}\bigl(\hat{\mathbf{X}}^{s+1}-\hat{\mathbf{X}}^{s}\bigr).
\end{align}
Here, \(\widetilde{\mathbf{X}}^{s+1}\) is the result of the gradient step, \(\hat{\mathbf{X}}^{s+1}\) is obtained by applying a sparse shrinkage followed by projection onto the nonnegative orthant, and \(\mathbf{X}^{s+1}\) denotes the accelerated iteration. The operator \(\soft_{\tau}(v)=\operatorname{sign}(v)\max(|v|-\tau,0)\) denotes element-wise soft-thresholding with threshold \(\tau\).
Once the inner loop has finished, its final iteration is used as the abundance estimate \(\mathbf{X}^k\) at stage \(k\). As a reference method in our experiments, a hard-constrained approach projects each abundance vector directly onto the simplex~\cite{heinz2001}.

\subsubsection{SVD-Guided DIP Residual Modeling ($\mathbf{S}$-Block).}
Following the update of the abundance matrix $\mathbf{X}^k$, the residual with respect to the fixed endmember matrix $\mathbf{A}$ is defined as
\begin{equation}
R_\mathbf{X}^k = \mathbf{Y}-\mathbf{A}\mathbf{X}^k .
\label{eq:rx_detailed}
\end{equation}
If the endmembers were perfectly known and the data strictly obeyed the linear mixing model, $R_\mathbf{X}^k$ would be dominated by small unstructured noise. In practice, however, the residual often displays organized patterns due to endmember mismatch, modeling artifacts, and other unmodeled effects. The aim of this stage is to make these structured discrepancies explicit, rather than absorbing them into the abundance estimates in the X-block. A residual update is expressed as
\begin{equation}
\mathbf{S}^k=\arg\min_\mathbf{S}\frac{1}{2}\norm{R_\mathbf{X}^k-\mathbf{S}}_F^2+\lambda_s\phi(\mathbf{S}),
\label{eq:s_sub_detailed}
\end{equation}
where $\phi(\mathbf{S})$ encodes the assumed structure prior to the residual branch. In sparse and low-rank configurations, $\phi(\mathbf{S})$ appears as an explicit regularizer; the simplest choice is an element-wise soft-thresholding rule, corresponding to the sparse prior $\phi(\mathbf{S})=\norm{\mathbf{S}}_1$:
\begin{equation}
\mathbf{S}^k =
\soft_{\lambda_s\hat{\sigma}^k}(R_\mathbf{X}^k),
\qquad
\hat{\sigma}^k=q_{0.99}(|R_\mathbf{X}^k|),
\label{eq:l1soft_detailed}
\end{equation}
with $q_{0.99}(\cdot)$ denoting the $99$th percentile of the absolute residual entries. This setup tests whether a purely sparse residual branch suffices, although mismatch-induced residuals for fixed endmembers are typically spatially and spectrally correlated, rather than strictly element-wise sparse.

To encode this correlated structure, we introduce a low-rank residual representation. We compute a truncated SVD of the transposed residual:
\begin{equation}
\mathbf{S}^k\approx R_\mathbf{X}^k
=U_{r_s}^k\Sigma_{r_s}^k(V_{r_s}^k)^{\top},
\label{eq:svd_residual_detailed}
\end{equation}
where $r_s$ denotes the residual guidance rank. Since $(R_\mathbf{X}^k)^{\top}\in\mathbb{R}^{n\times \ell}$ treats pixels
as samples and spectral bands as feature dimensions, the leading singular directions capture the main residual patterns shared across pixels. This low-rank subspace strategy is motivated by recent hyperspectral denoising work that combines subspace decomposition with neural priors~\cite{hysudeep2021}.

Classical priors such as sparsity or low-rankness explain only part of the residual structure and are restricted to element-wise or linear models. In contrast, the default SVD-guided DIP branch---denoted learnable reparameterization lowrank\_dip in the experiments---imposes a residual prior indirectly rather than penalizing $\mathbf{S}$ explicitly. Instead of solving a direct $\mathbf{S}$-subproblem, we reparameterize $\mathbf{S}$ via a neural mapping $f_\theta$, constraining $\mathbf{S}$ to the image of $f_\theta$; the optimization over $\theta$ then implicitly induces the structural prior. Concretely, the residual generator is given by
\begin{equation*}
\mathbf{S}_{\theta}^k = f_{\theta}(Z^k)^{\top},\; Z^k=(z_i^k,\dots,z_n^k)^{\top} = U_{r_s}^k\Sigma_{r_s}^k
\in\mathbb{R}^{n\times r_s},
\label{eq:dip_param_detailed}
\end{equation*}
where a shared nonlinear lifting network $f_{\theta}:\mathbb{R}^{r_s}\rightarrow\mathbb{R}^{\ell}$ maps each $z_i^k$ to a $\ell$-dimensional residual spectrum $f_{\theta}(z_i^k)$.
Using decomposition-derived coordinates as input to a coordinate-based MLP follows the implicit neural representation strategy recently adopted for hyperspectral image restoration~\cite{lrtsinr2025}.
This constrains residuals to follow the dominant SVD modes, while allowing nonlinear variations within the induced subspace. The parameters $\theta$ are obtained by solving
\begin{equation}
\theta^{k,*}
=
\arg\min_{\theta}
\frac{1}{2}
\norm{
R_\mathbf{X}^k - f_{\theta}(Z^k)^{\top}
}_F^2 .
\label{eq:dip_fitting_detailed}
\end{equation}

In practice, $f_{\theta}$ is a compact per-pixel two-layer MLP with architecture $r_s\rightarrow 32\rightarrow \ell$ and ReLU activations. For the Urban dataset with $r_s=3$ and $\ell=162$, this yields about $5.5\times10^3$ parameters per residual instance. Each stage is initialized randomly and trained with Adam (learning rate $10^{-3}$) for up to $1500$ iterations. Unless otherwise stated, we set $r_s=3$ and $T=2$.

\section{Experiments}
\label{sec:experiments}
\subsection{Experimental Setup}
We assess the proposed method---in which the fixed endmember matrix $\mathbf{A}$ is obtained in the A-block---on three benchmark hyperspectral datasets{~\cite{zhu2017hsudata}} with different scene properties: Samson ($156$ bands, $95\times95$ pixels, $r=3$), Urban ($162$ bands, $307\times307$ pixels, $r=4$), and Jasper Ridge ($198$ bands, $100\times100$ pixels, $r=4$). Unless noted otherwise, the $\mathbf{X}$-block is solved with FISTA using $\lambda_x = 10^{-4}$, $\mu = 10^{-1}$, and at most $2000$ iterations. In the $\mathbf{S}$-block, the $\ell_1$-soft branch adopts $\lambda_s = 0.2$ after residual normalization by $q_{0.99}(|R|)$, while both lowrank\_r3 and lowrank\_dip are governed by a residual rank of $r_s = 3$. The default lowrank\_dip branch employs a two-layer MLP with $32$ hidden units, trained with Adam for $1500$ iterations at a learning rate of $10^{-3}$. This configuration is applied uniformly to all datasets, with no dataset-specific tuning.

\begin{sloppypar}
For endmember extraction, we report the mean Spectral Angle Mapper (SAM, degrees)~\cite{bioucas2012overview} after obtaining estimated and reference endmembers. For fixed-prior unmixing, we quantify abundance accuracy by $x_{\rm rmse}$ and decomposition fidelity by reconstruction SAM and the relative joint error,
\end{sloppypar}
\begin{equation}
x_{\rm rmse}=\sqrt{\tfrac{1}{rN}\norm{\mathbf{X}-\hat{\mathbf{X}}}_F^2},
\qquad
\mathrm{Joint}=\frac{\norm{\mathbf{Y}-A\hat{\mathbf{X}}-\hat{\mathbf{S}}}_F}{\norm{\mathbf{Y}}_F},
\label{eq:x_rmse}
\end{equation}
where $\mathbf{X},\hat{\mathbf{X}}$ are the ground-truth and estimated abundances~\cite{iordache2011} and $\hat{\mathbf{S}}=0$ for methods without a residual component. The soft-versus-hard comparison reports the abundance sum violation $\tfrac{1}{N}\sum_{i=1}^{N}\big|1-\mathbf{1}^\top\bm{x}_i\big|$
to measure deviations from the sum-to-one constraint.

\begin{sloppypar}
\textbf{Reproducibility.} All main runs use one fixed N-FINDR realization (Table~\ref{tab:a_compare}) and a fixed random seed ($0$; seeds $0$--$2$ are used only for the stability analysis in Fig.~\ref{fig:rank_efficiency}b), with no ground-truth-based selection. Since $\lambda_x$ and $\mu$ are fixed across scenes with different scales, quantitative comparisons are made within each dataset rather than across datasets.
\end{sloppypar}

\subsection{Reliability of Fixed Endmember Priors}
We first evaluate the A-block to establish a reasonable but imperfect frozen prior for the downstream fixed-prior experiments, rather than to select an oracle endmember extractor.

Table~\ref{tab:a_compare} indicates that no single extractor dominates across all datasets. ATGP attains the lowest SAM on Urban, while N-FINDR achieves the best performance on Samson and Jasper and remains competitive with ATGP on Urban. We therefore adopt N-FINDR as the default A-block, not as a universally optimal method, but because it offers the most balanced fixed prior over the three datasets. This distinction matters: the resulting prior remains imperfect---especially on Urban---and the downstream hyperspectral unmixing solver is explicitly designed to handle such fixed-prior mismatch.

Once $\mathbf{A}$ is fixed, the subsequent experiments focus on isolating the influence of modeling choices in $\mathbf{X}$ and $\mathbf{S}$ under the default N-FINDR prior. VCA, ATGP, and FIPPI are each run through the complete pipeline in the prior-mismatch stress test (Fig.~\ref{fig:prior_quality}). In that test, each prior is re-extracted independently using the corresponding extractor settings, so the resulting per-extractor SAM values may differ from those reported in Table~\ref{tab:a_compare}.

\begin{table}[!tbp]
\centering
\caption{Comparison of endmember extractors using SAM (deg).}
\label{tab:a_compare}
\setlength{\tabcolsep}{2.mm}{\small
\begin{tabular*}{0.95\hsize}{@{}@{\extracolsep{\fill}}lccc@{}}
\toprule
Method & Samson & Urban & Jasper \\
\midrule
VCA \cite{nascimento2005vca}     & 4.89             & 25.30            & 22.09            \\
N-FINDR \cite{winter1999nfindr} & \textbf{4.02}   & 25.57            & \textbf{12.17}   \\
ATGP \cite{ren2003atgp}    & 21.99           & \textbf{24.82}   & 18.50            \\
FIPPI \cite{chang2006fippi}   & 20.45           & 31.01            & 25.31            \\
\bottomrule
\end{tabular*}}
\end{table}

\subsection{Soft Relaxation Versus Hard Simplex}
Under an imperfect prior, we compare a soft sum-to-one constraint (soft ASC) with hard simplex projection, and examine where the relaxation is applied.

Table~\ref{tab:soft_vs_hard} isolates the effect of the abundance constraint. Soft abundance relaxation produces lower abundance RMSE and reduced joint error. Although the hard-simplex solver enforces exact sum-to-one, strict simplex feasibility of the abundance vector $\bm{x}_i$ does not necessarily yield better decompositions when the prior is inaccurate.

\begin{figure}[!tbp]
\centering
\includegraphics[width=0.8\textwidth]{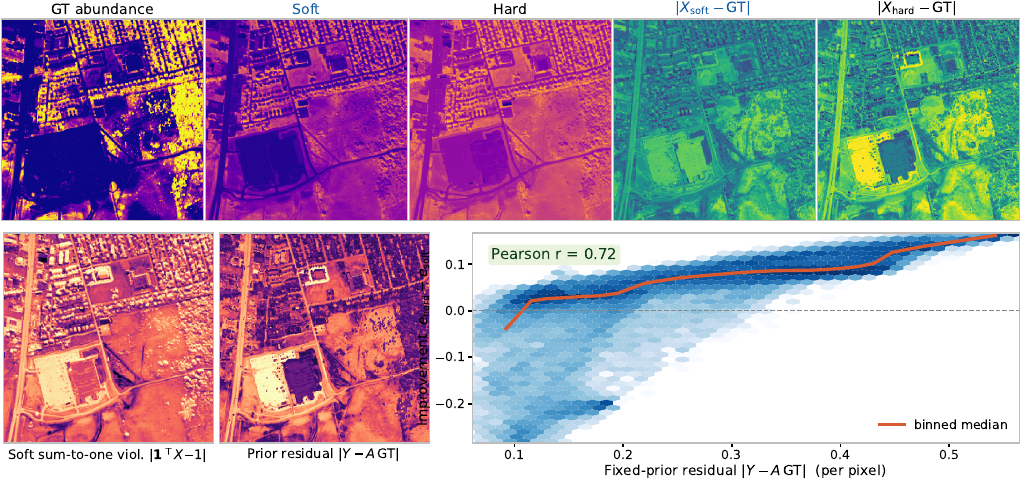}
\caption{ Soft-ASC vs hard-simplex spatial mechanism on Urban (fixed N-FINDR prior). Top: ground-truth, soft, and hard abundance maps with per-pixel error maps. Bottom: the soft sum-to-one violation map and the fixed-prior residual $|\mathbf{Y}-\mathbf{A}\,\mathrm{GT}|$ co-localize, and the per-pixel scatter shows the soft-over-hard improvement $e_{\rm hard}-e_{\rm soft}$ growing with the prior residual.}
\label{fig:soft_hard_mech}
\end{figure}

 Fig.~\ref{fig:soft_hard_mech} makes this spatial mechanism explicit on Urban: the soft sum-to-one violation concentrates where the fixed-prior residual $|\mathbf{Y}-\mathbf{A}\,\mathrm{GT}|$ is large, and a per-pixel analysis shows that the soft-over-hard error improvement $e_{\rm hard}-e_{\rm soft}$ grows with that prior residual (Pearson $r=0.72$; median improvement is about $5\times$ larger in the most unreliable quartile than in the most reliable one). Soft relaxation therefore loosens the constraint selectively in unreliable regions rather than uniformly.

\begin{table}[!tbp]
\centering
\caption{Comparison between soft abundance relaxation and hard simplex projection under the same fixed N-FINDR prior.}
\label{tab:soft_vs_hard}
\setlength{\tabcolsep}{1.mm}{\small
\begin{tabular*}{0.95\hsize}{@{}@{\extracolsep{\fill}}lcccccc@{}}
\toprule
\multirow{2}{*}{Dataset} & \multicolumn{2}{c}{RMSE$\downarrow$} & \multicolumn{2}{c}{Sum viol.$\downarrow$} & \multicolumn{2}{c}{Joint$\downarrow$} \\
\cmidrule(lr){2-3}\cmidrule(lr){4-5}\cmidrule(lr){6-7}
 & Soft & Hard & Soft & Hard & Soft & Hard \\
\midrule
Urban & \textbf{0.3239} & 0.3895 & 0.5504 & \textbf{0.0000} & \textbf{0.1002} & 0.4965 \\
Samson & \textbf{0.4057} & 0.4470 & 0.3134 & \textbf{0.0000} & \textbf{0.0357} & 0.0525 \\
Jasper & \textbf{0.4212} & 0.4612 & 0.3383 & \textbf{0.0000} & \textbf{0.0900} & 0.1979 \\
\bottomrule
\end{tabular*}}
\end{table}

\subsection{Ablation Study for Hyperspectral Unmixing Enhancement}
To isolate the source of the improvement, we vary the S-block across residual sparsity, low-rank structure, and the SVD-guided DIP branch, and further include a generic Joint-DIP control.

\begin{figure}[!tbp]
\centering
\includegraphics[width=0.95\textwidth]{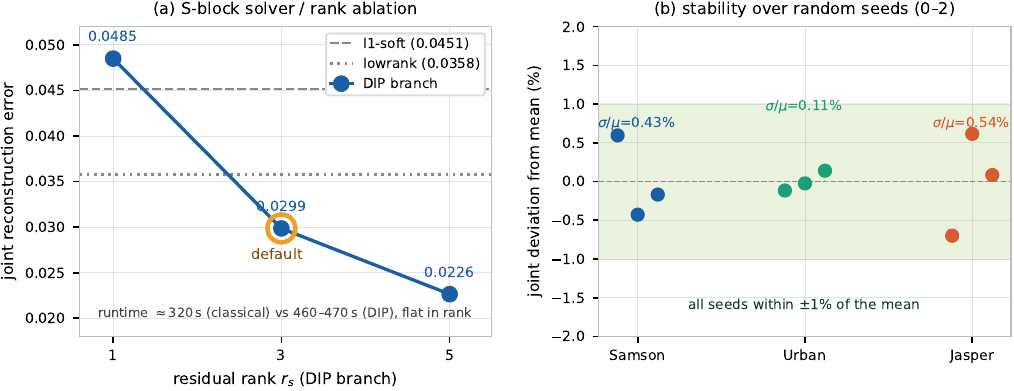}
\caption{ \emph{(a)} S-block solver/rank ablation on Urban under the fixed N-FINDR prior: joint error versus residual rank for the DIP branch, with non-DIP \texttt{l1-soft} and low-rank S-blocks shown as references. \emph{(b)} stability of the joint error over three random DIP seeds; all runs stay within $\pm1\%$ of the per-dataset mean.}
\label{fig:rank_efficiency}
\end{figure}

As shown in Table~\ref{tab:ablation_results}, each successive refinement of the S-block produces consistent gains in SAM and joint error across all three datasets, with $x_{\rm rmse}$ remaining essentially unchanged. These results suggest that the gain comes mainly from the SVD-guided DIP residual branch rather than from simply adding an arbitrary residual term. Furthermore,
 the DIP branch substantially reduces the joint error compared with the non-DIP \texttt{l1-soft} and analytic low-rank S-blocks (Fig.~\ref{fig:rank_efficiency}). Because the runtime varies little across residual ranks in this setting, we use $(T,r_s)=(2,3)$ as a globally fixed default rather than a tuned optimum, and report $r_s{=}5$ only as an exploratory higher-capacity variant.

 Two controls support the design. A frozen-$\mathbf{A}$ Joint-DIP baseline, which performs joint optimization without the stage-wise SVD-guided routing, attains slightly lower abundance RMSE in some cases but worse reconstruction SAM and joint error than the staged design on all three datasets (Joint-DIP rows, Table~\ref{tab:ablation_results}). This indicates that the improvement is not simply due to adding a generic DIP module, but to the proposed stage-wise routing between $\mathbf{X}$ and $\mathbf{S}$. In addition, across three DIP seeds the joint error stays within $\pm1\%$ of its mean (Fig.~\ref{fig:rank_efficiency}b).

\begin{table}[!tbp]
\centering
\caption{Hyperspectral unmixing enhancement path across the three datasets. Bold marks the best value along the staged enhancement path (X-only $\to$ \texttt{l1-soft} $\to$ \texttt{lowrank\_r3} $\to$ \texttt{lowrank\_dip}); the Joint-DIP row is a separate control and is excluded from the bold comparison.}
\label{tab:ablation_results}
\setlength{\tabcolsep}{2.mm}{\small
\renewcommand{\arraystretch}{0.92}
\setlength{\aboverulesep}{0.25ex}
\setlength{\belowrulesep}{0.35ex}
\begin{tabular*}{0.95\hsize}{@{}@{\extracolsep{\fill}}llccc@{}}
\toprule
Data & Variant & RMSE$\downarrow$ & SAM$\downarrow$ (deg) & Joint$\downarrow$ \\
\midrule
\multirow{5}{*}{Samson} & X-only & 0.4057 & 2.7589 & 0.0357 \\
 & l1-soft & \textbf{0.4051} & 1.7629 & 0.0166 \\
 & lowrank\_r3 & 0.4056 & 1.3555 & 0.0129 \\
 & lowrank\_dip (default) & 0.4056 & \textbf{1.0305} & \textbf{0.0107} \\[0.3ex]
 & Joint-DIP & 0.3932 & 2.3484 & 0.0342 \\
\midrule
\multirow{5}{*}{Urban} & X-only & 0.3239 & 5.9915 & 0.1002 \\
 & l1-soft & 0.3236 & 3.3214 & 0.0462 \\
 & lowrank\_r3 & \textbf{0.3235} & 2.4473 & 0.0366 \\
 & lowrank\_dip (default) & 0.3236 & \textbf{1.9206} & \textbf{0.0295} \\[0.3ex]
 & Joint-DIP & 0.3113 & 2.3699 & 0.0395 \\
\midrule
\multirow{5}{*}{Jasper} & X-only & 0.4212 & 12.3878 & 0.0900 \\
 & l1-soft & 0.4203 & 6.1206 & 0.0411 \\
 & lowrank\_r3 & \textbf{0.4177} & 3.1454 & 0.0316 \\
 & lowrank\_dip (default) & 0.4188 & \textbf{2.5208} & \textbf{0.0251} \\[0.3ex]
 & Joint-DIP & 0.3523 & 3.3525 & 0.0379 \\
\bottomrule
\end{tabular*}}
\end{table}

\subsection{Qualitative Residual Refinement and Abundance Stability}
We now inspect the decomposition visually, checking that $\mathbf{S}$ absorbs structured spatial--spectral mismatch while the abundance maps stay stable.

\begin{figure}[!tbp]
\centering
\includegraphics[width=\textwidth]{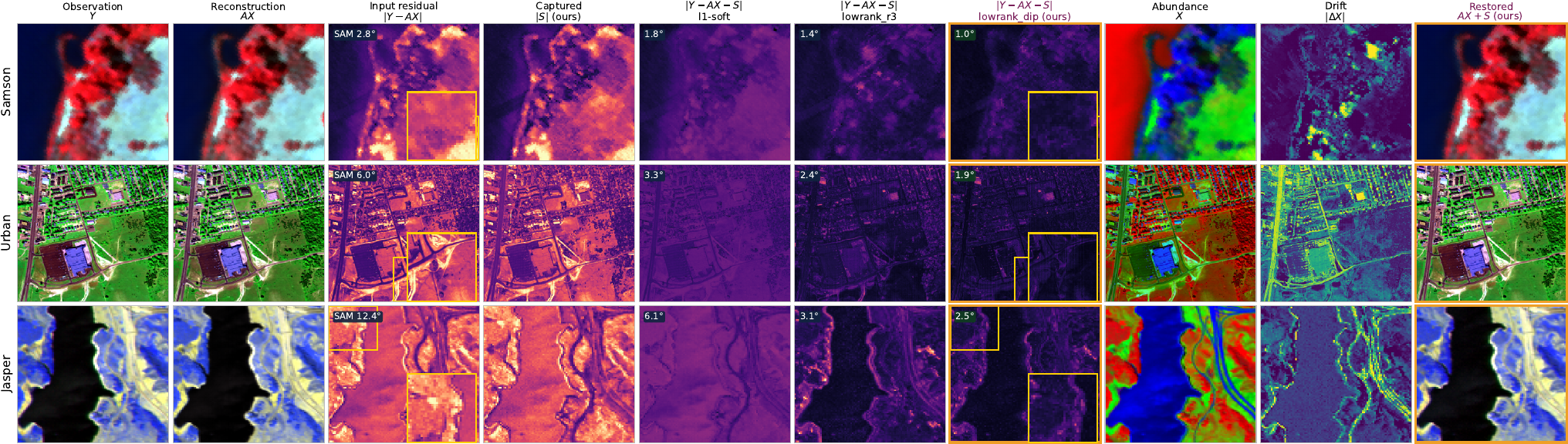}
\caption{ S-block residual suppression on Samson, Urban, and Jasper under the fixed N-FINDR prior. Columns show the input image, $\mathbf{A}\mathbf{X}$, the input/captured/final residuals for the different S-blocks (\texttt{l1-soft}, \texttt{lowrank\_r3}, \texttt{lowrank\_dip}), abundance $\mathbf{X}$, drift $|\Delta\mathbf{X}|$, and restored $\mathbf{A}\mathbf{X}+\mathbf{S}$. Residual maps share a per-row scale; badges report reconstruction SAM.}
\label{fig:residual_routing_3ds}
\end{figure}

\begin{figure}[!tbp]
\centering
\includegraphics[width=\textwidth]{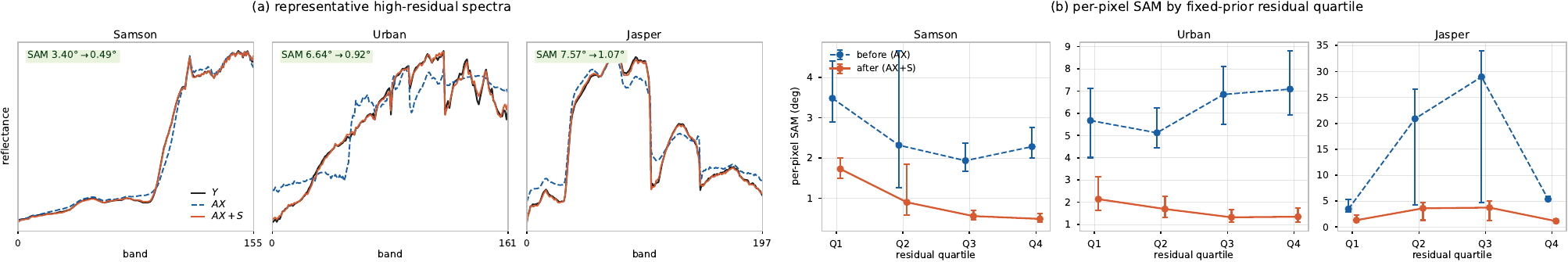}
\caption{ Spectral evidence of residual refinement. \emph{(a)} for one high-residual pixel per dataset, $\mathbf{A}\mathbf{X}+\mathbf{S}$ better matches $\mathbf{Y}$ and reduces per-pixel SAM. \emph{(b)} across all pixels grouped by fixed-prior residual quartiles, $\mathbf{S}$ consistently lowers the median SAM, with larger gains in higher-residual regions.}
\label{fig:spectral_evidence}
\end{figure}

 We first examine how structured mismatch is transferred from the reconstruction residual into the $\mathbf{S}$ branch: progressively richer residual models (\texttt{l1-soft}, \texttt{lowrank\_r3}, \texttt{lowrank\_dip}) leave a weaker final residual (Fig.~\ref{fig:residual_routing_3ds}), with the default \texttt{lowrank\_dip} attaining the lowest reconstruction SAM on every dataset. Fig.~\ref{fig:spectral_evidence} confirms the spectral role of $\mathbf{S}$: $\mathbf{A}\mathbf{X}+\mathbf{S}$ aligns with $\mathbf{Y}$ at high-residual pixels, and across residual quartiles $\mathbf{S}$ consistently lowers the median SAM, especially in high-mismatch regions. This supports the interpretation that $\mathbf{S}$ accounts for structured mismatch while inducing only small abundance drift.

Overall, routing structured mismatch into $\mathbf{S}$ leaves $\mathbf{X}$ stable: $\Delta x_{\rm rmse}$ stays within $0.003$ and $\Delta\mathbf{X}_{\rm rms}$ within $0.022$ across all datasets.


\subsection{Comparisons With Other Methods}
With the A-block fixed and soft relaxation verified, we assess the full I-HyperSU against classical solvers under the same frozen-$\mathbf{A}$, no-extra-training protocol.

\begin{table}[!tbp]
\centering
\caption{Overall comparison under the same fixed-$\mathbf{A}$ protocol. The proposed I-HyperSU denotes the default lowrank\_dip configuration.}
\label{tab:main_results}
\setlength{\tabcolsep}{2.mm}{\small\renewcommand{\arraystretch}{0.90}
\begin{tabular*}{0.95\hsize}{@{}@{\extracolsep{\fill}}llccc@{}}
\toprule
Data & Method & RMSE$\downarrow$ & SAM$\downarrow$ (deg) & Joint$\downarrow$ \\
\midrule
\multirow{4}{*}{Samson} & UCLS\cite{keshava2002} & 0.4117 & 2.6979 & 0.0351 \\
 & FCLS\cite{heinz2001} & 0.4470 & 4.4569 & 0.0525 \\
 & NNLS\cite{lawson1995} & 0.4057 & 2.7589 & 0.0357 \\
 & I-HyperSU (ours) & \textbf{0.4056} & \textbf{1.0305} & \textbf{0.0107} \\
\midrule
\multirow{4}{*}{Urban} & UCLS\cite{keshava2002} & 0.3240 & 5.1824 & 0.0894 \\
 & FCLS\cite{heinz2001} & 0.3895 & 21.4605 & 0.4965 \\
 & NNLS\cite{lawson1995} & 0.3268 & 5.7716 & 0.0993 \\
 & I-HyperSU (ours) & \textbf{0.3236} & \textbf{1.9206} & \textbf{0.0295} \\
\midrule
\multirow{4}{*}{Jasper} & UCLS\cite{keshava2002} & 0.4758 & 7.7346 & 0.0656 \\
 & FCLS\cite{heinz2001} & 0.4612 & 13.7153 & 0.1979 \\
 & NNLS\cite{lawson1995} & 0.4212 & 12.3812 & 0.0898 \\
 & I-HyperSU (ours) & \textbf{0.4188} & \textbf{2.5208} & \textbf{0.0251} \\
\bottomrule
\end{tabular*}}
\end{table}

Table~\ref{tab:main_results} summarizes the primary comparison. Across all three datasets, the proposed I-HyperSU attains the best reconstruction SAM and joint error, while preserving an abundance RMSE comparable to the strongest classical baseline. This aligns with the method's design goal: the proposed I-HyperSU prioritizes decomposition consistency under fixed priors rather than maximizing abundance prediction under an oracle model.

To ensure a fair downstream comparison in the frozen-prior regime, external baselines are limited to solvers that neither update $\mathbf{A}$ nor rely on extra training data. We also tested SUnSAL under the same fixed-$\mathbf{A}$ setting; its results were nearly identical to NNLS under the adopted parameters, so we omit it from the main table for compactness. Relative to UCLS, the proposed I-HyperSU reduces joint error by roughly $69.5\%$, $67.0\%$, and $61.7\%$ on Samson, Urban, and Jasper, respectively, while maintaining competitive abundance RMSE. These improvements demonstrate that explicit residual refinement enhances decomposition consistency when endmember priors are fixed.

\subsection{Synthetic Validation with a Known Residual}
On observed data, adding $\mathbf{S}$ lowers the reconstruction error by construction, so a good fit alone does not prove that $\mathbf{S}$ captures meaningful mismatch. We therefore turn to a synthetic setting to test whether $\mathbf{S}$ recovers genuine out-of-model structure rather than merely fitting the observation residual.

We build $\mathbf{Y}=\max(\mathbf{A}_{\rm gt}\mathbf{X}_{\rm gt}+\mathbf{S}_0+\bm{\eta},0)$ with a \emph{known} non-negative residual $\mathbf{S}_0$ (stray-light/haze + striping) on the Jasper reference. To make the test diagnostic, $\mathbf{S}_0$ is constructed as a rank-$5$ component with a large portion outside $\mathrm{span}(\mathbf{A}_{\rm gt})$, so that it cannot be fully absorbed by abundance changes. We then sweep its strength $\|\mathbf{S}_0\|/\|\mathbf{Y}_0\|\in\{0,0.1,0.2,0.3\}$ ($\mathbf{S}_0{=}0$ is a control), running the \emph{unmodified} pipeline with $\mathbf{A}=\mathbf{A}_{\rm gt}$ fixed.

\begin{figure}[!tbp]
\centering
\includegraphics[width=\textwidth]{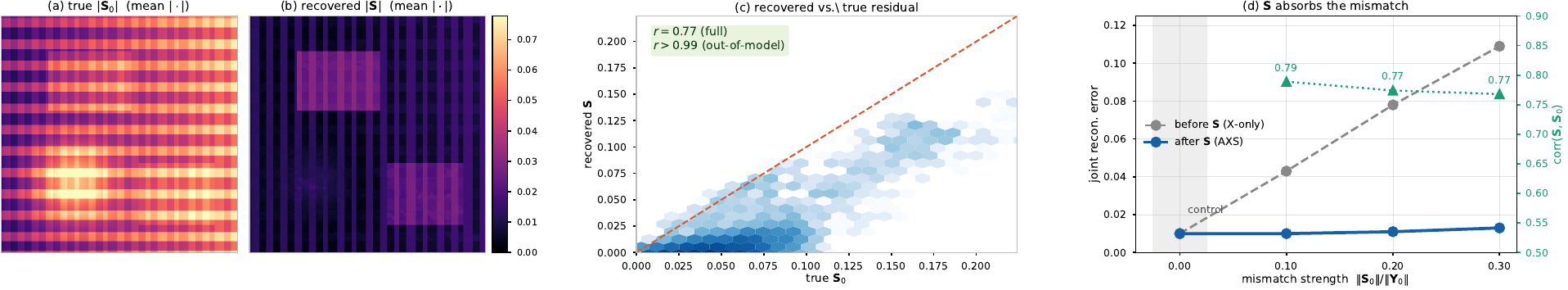}
\caption{ Synthetic validation with a known non-negative residual $\mathbf{S}_0$ (stray-light/haze + striping; rank $5$; partly out of $\mathrm{span}(\mathbf{A})$; true endmembers fixed). \emph{(a,b)} true $|\mathbf{S}_0|$ vs.\ recovered $|\mathbf{S}|$ (strength $0.20$). \emph{(c)} recovered vs.\ true residual. \emph{(d)} joint error before vs.\ after $\mathbf{S}$ across mismatch strengths, with corr$(\mathbf{S},\mathbf{S}_0)$ and the $\mathbf{S}_0{=}0$ control.}
\label{fig:synthetic_S}
\end{figure}

Three findings (Fig.~\ref{fig:synthetic_S}) show that $\mathbf{S}$ captures real mismatch. (i) $\mathbf{S}$ correlates with the true $\mathbf{S}_0$ (Pearson $r{\approx}0.77$), rising to $r\!>\!0.99$ on the out-of-subspace component that abundance changes cannot explain. (ii) Routing the residual into $\mathbf{S}$ substantially reduces the joint error at every mismatch strength. (iii) The $\mathbf{S}_0{=}0$ control confirms that $\mathbf{S}$ does not fabricate structure, with abundance RMSE essentially unchanged. Recovery is partial in magnitude because the in-$\mathrm{span}(\mathbf{A})$ component can be shared with $\mathbf{A}\mathbf{X}$, but the ground-truth correlation and zero-mismatch control distinguish genuine recovery from residual fitting.

\subsection{Prior-Mismatch Stress Test}
\begin{figure}[!htbp]
\centering
\includegraphics[width=0.72\textwidth]{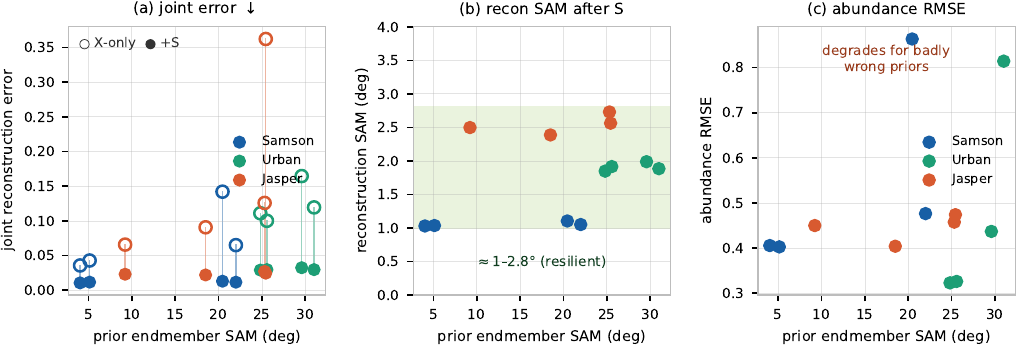}
\caption{ Prior-mismatch stress test over four extractors (N-FINDR, VCA, ATGP, FIPPI) on the three datasets. \emph{(a)} X-only joint error ($\circ$) grows with prior SAM while joint error with $\mathbf{S}$ ($\bullet$) stays low. \emph{(b)} reconstruction SAM after $\mathbf{S}$ stays low across extractors. \emph{(c)} abundance RMSE instead degrades for badly wrong priors.}
\label{fig:prior_quality}
\end{figure}

To map the operating regime of I-HyperSU, we repeat the full pipeline under four extractors of differing quality (N-FINDR, VCA, ATGP, and FIPPI). The residual branch is \emph{robust in reconstruction}: with $\mathbf{S}$, joint error and reconstruction SAM remain low across extractors (Fig.~\ref{fig:prior_quality}a,b). Abundance accuracy, however, is \emph{not} immune and degrades for severely wrong priors (FIPPI; Fig.~\ref{fig:prior_quality}c). Thus, $\mathbf{S}$ preserves reconstruction consistency but cannot restore abundance identifiability once the endmembers are badly inaccurate. This indicates that I-HyperSU is reconstruction-resilient under imperfect priors, but not prior-independent: accurate abundance recovery still requires reasonably reliable endmembers.

\FloatBarrier
\section{Conclusion}
\begin{sloppypar}
This paper reformulated fixed-prior hyperspectral unmixing as an explicit
mismatch-optimizing task and introduced an I-HyperSU framework that
separates the fixed endmember prior, soft-constrained abundance estimation,
and structured residual modeling into three coupled but inspectable components.
Experiments on Samson, Urban, and Jasper Ridge show that soft abundance
relaxation consistently outperforms hard simplex projection under fixed and
imperfect endmember priors, and that the default low-rank-guided DIP residual branch
reduces joint reconstruction error by $61.7\%$--$69.5\%$ relative to UCLS with negligible impact on abundance
RMSE under the default N-FINDR prior. Synthetic and prior-mismatch experiments further show that $\mathbf{S}$ captures genuine out-of-model mismatch ($r\!\approx\!0.77$) and improves reconstruction resilience, while accurate abundance recovery still requires a reasonable prior. By separating material abundances from structured spectral mismatch, I-HyperSU provides an interpretable scene-decomposition view of hyperspectral perception under fixed endmember priors.
\end{sloppypar}

This study focuses on three standard datasets and a frozen-$\mathbf{A}$ configuration. Extending the framework to variability-aware baselines and end-to-end deep methods under a unified frozen-$\mathbf{A}$ protocol, and developing reliability-aware refinement that adapts to the quality of the prior, are left for future work.

\begin{credits}
\subsubsection{\discintname}
The authors have no competing interests to declare that are relevant to the
content of this article.
\end{credits}

%
\bibliographystyle{splncs04}
\bibliography{references}
\end{document}